%% file: main.tex
\documentclass[letterpaper, 10 pt, conference]{ieeeconf}  
\IEEEoverridecommandlockouts                              
\usepackage[usenames, dvipsnames]{color}
\usepackage[noadjust]{cite}
\usepackage{booktabs}
\usepackage{graphicx}
\usepackage{subcaption}
\usepackage{float}
\usepackage{amsmath}
\usepackage{algorithmicx}
\usepackage{multirow}
\usepackage{mathrsfs}
\graphicspath{{images}{../images/}}
\usepackage{graphicx}
\usepackage[utf8]{inputenc}
\usepackage{amsmath}
\usepackage[hidelinks]{hyperref}
\usepackage{stix,bbding,pifont,utfsym,fontawesome}

\usepackage{soul}
\usepackage{siunitx}
\usepackage[export]{adjustbox}
\usepackage{multicol}
\usepackage{tikz}
\usepackage[symbol]{footmisc}

\title{\LARGE \bf
Flow-guided Motion Prediction with Semantics and \\Dynamic Occupancy Grid Maps
}
	
\author{Rabbia Asghar, Wenqian Liu, Lukas Rummelhard, Anne Spalanzani, Christian Laugier\\
{\small Univ. Grenoble Alpes, Inria, 38000 Grenoble, France, email:  FirstName.LastName@inria.fr}%
}

\begin{document}
\setstcolor{red}

\maketitle
\thispagestyle{empty}
\pagestyle{empty}
\setlength{\belowdisplayskip}{2pt}
\widowpenalty10000
\clubpenalty10000
\addtolength{\abovedisplayskip}{-5pt}

\begin{abstract}
Accurate prediction of driving scenes is essential for road safety and autonomous driving. Occupancy Grid Maps (OGMs) are commonly employed for scene prediction due to their structured spatial representation, flexibility across sensor modalities and integration of uncertainty. Recent studies have successfully combined OGMs with deep learning methods to predict the evolution of scene and learn complex behaviours. These methods, however, do not consider prediction of flow or velocity vectors in the scene. In this work, we propose a novel multi-task framework that leverages dynamic OGMs and semantic information to predict both future vehicle semantic grids and the future flow of the scene. This incorporation of semantic flow not only offers intermediate scene features but also enables the generation of warped semantic grids. Evaluation on the real-world NuScenes dataset demonstrates improved prediction capabilities and enhanced ability of the model to retain dynamic vehicles within the scene. 
\end{abstract}

\input{sections/introduction}
\input{sections/related_work}
\input{sections/approach}
\input{sections/experiments}
\input{sections/results.tex}
\input{sections/conclusion}




%
%

\bibliographystyle{IEEEtran}
\bibliography{main}%

\end{document}

%% file: sections/introduction.tex
\section{INTRODUCTION} \label{sec:introduction}

Predicting driving scenes holds significant importance in enhancing road safety, optimizing traffic flow, and advancing autonomous driving technologies. By accurately anticipating various elements within a driving environment such as pedestrian movements, and vehicle trajectories,
predictive models enable proactive decision-making for both human drivers and autonomous systems. This foresight aids in mitigating risks, preventing accidents, and ultimately saving lives. 

Dynamic Occupancy Grid Maps (DOGMs) represents static and dynamic elements within the environment in a bird's-eye-view (BEV) grid. One of the key advantages lies in their flexibility regarding sensor dependency, as they can be generated from various types of sensors including Lidar, radars, and cameras. In our research, we leverage probabilistic DOGMs \cite{lukas22},
which further enhances predictive capabilities by incorporating uncertainty estimation. Moreover, by integrating deep learning techniques with such probabilistic DOGMs, we can effectively learn complex interactions within the driving environment, tackle the intrinsic spatiotemporal problems, and lead to more robust and precise predictions.

In the domain of autonomous vehicles, Occupancy Grid Map (OGM) predictions are often ego-centric (\cite{Toyungyernsub2022}, \cite{mann2022predicting}), meaning they are fixed relative to the ego-vehicle frame. Dependent on the motion of the ego-vehicle, the future predicted grids suffer from the disappearance of static and dynamic scenes over time and loss of scene structure, especially during turns. 
To address these shortcoming, Asghar \textit{et al.} \cite{asghar2022allo} proposed making allo-centric DOGM predictions by representing the scene evolution in a fixed reference frame. This approach successfully preserves scene integrity independent of ego-vehicle motion and focuses on the behavior and motion prediction of dynamic agents in the scene.

Asghar \textit{et al.} \cite{asghar2023vehicle} extended this work to consider availability of semantic information with the input and to predict agents motion in the scene as sequential BEV semantic grids. The framework demonstrated improved abilities to predict dynamic agents as well as the possibility to evaluate with the ground truth.

In this work, we build further on the existing approaches to formulate a multi-task framework by learning the flow of the scene to improve future semantic grids. 
For our input we consider both occupancy probabilities and estimated velocities of the dynamic scene components from DOGMs, along with the semantic information from the camera images. Our future predictions are allo-centric, fixed with respect to the latest ego position. 
We generate predictions for both the future semantic grids and the flows.
The predicted flows also serve as intermediate features, capturing to the spatial and temporal changes in the semantic information within the scene. By forecasting the flows, we anticipate how the semantic content of the scene will evolve over time. Subsequently, we iteratively warp the current semantic grid with the future flows to obtain the sequence of future semantic grids.  A brief overview of this work can be accessed online\footnote[2]{https://youtu.be/rhyC2Xo0B-g}.

The main contribution of this paper is a novel multi-task framework that combines DOGM and semantics to predict multi-step vehicle semantic grids, as well as the multi-step flow of the complete scene. Evaluation on real-world NuScenes dataset \cite{nuscenes2019} shows how the proposed framework improves prediction capabilities and learns to recognize vehicle behaviors within the scene without prior maps.

\begin{figure*}[ht]
	\centering
	    \includegraphics[trim={0 0 0 0},clip,width=1.9\columnwidth]{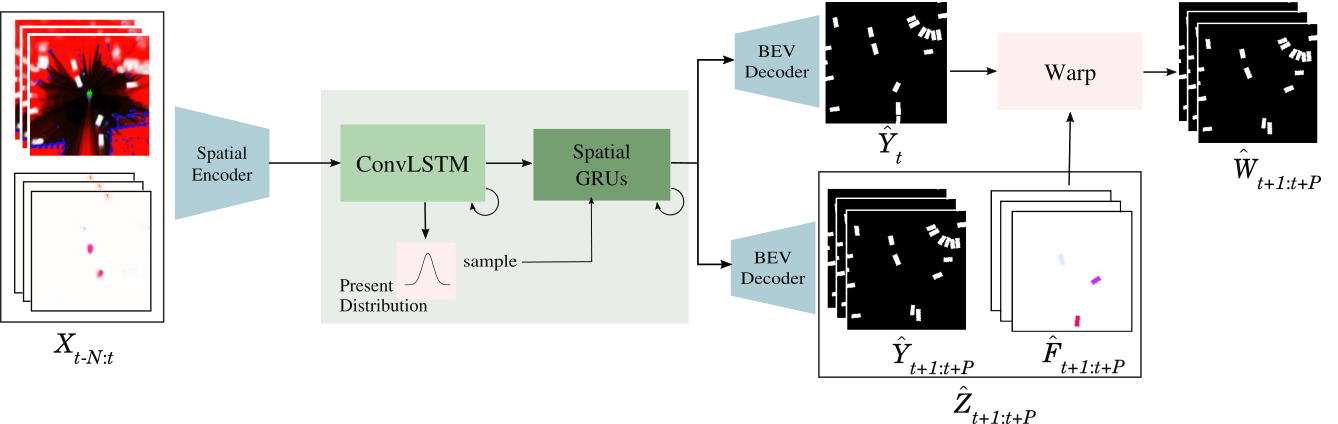}
\caption{\small An overview of our proposed network. Our network utilizes a sequence of DOGMs, occupancy state grids, associated velocity grids, and semantic grids $X_{t-N:t}$ as input to capture the scene evolution. Subsequently, it predicts a sequence of future vehicle semantic grids $\hat{Y}_{t+1:t+P}$ and scene flows $\hat{F}_{t+1:t+P}$.} 
\label{fig:overview}
\vspace{-0.5cm}
\end{figure*}

%% file: sections/related_work.tex
\section{Related Work} \label{sec:related_work}

\subsection{Occupancy Grid Prediction}

In the state-of-the-art literature, the challenge of predicting ego-centric future OGM (Occupancy Grid Map) involves incorporating spatio-temporal deep-learning methods (\cite{8500567},  \cite{Dequaire2017}, \cite{mohajerin2019multi}, \cite{lange2020attention}, ).
Asghar \textit{et al.} \cite{asghar2022allo} proposed a novel approach for future DOGM prediction within a fixed reference frame, preserving scene integrity particularly during ego-vehicle maneuvers.

Studies by Toyungyernsub \textit{et al.} \cite{Toyungyernsub2022} and Schreiber \textit{et al.} \cite{schreiber2019long} proposed separate predictions of the static and dynamic scene, yet they do not predict velocities or scene flows. 
 Mann \textit{et al.} \cite{mann2022predicting} enhanced DOGMs with semantic vehicle labels, while \cite{asghar2023vehicle} integrated DOGMs with additional map and vehicle semantics, thus improving prediction capabilities and accommodating diverse future scenarios through a probability distributions module. 

In this paper, contrary to the conventional OGM prediction methods, we forecast both the vehicle occupancy grids and the dynamics of the entire scene. We carry forward the work of Asghar \textit{et al.} \cite{asghar2023vehicle}, incorporate velocity information, predict the scene flows in the output, generate warped semantic grids, and demonstrate how these enhancements can improve the prediction capabilities of vehicles in the scene.

\subsection{Motion Forecasting via Occupancy Grids}

In the realm of motion forecasting, the literature tends to predominantly focus on approaches utilizing detection, tracking, and trajectory prediction of agents. Nevertheless, recent efforts have emerged to explore the efficacy of occupancy grids for motion forecasting, albeit to a lesser extent. Notably, Rules of the Road \cite{hong2019rules} presented a comparative analysis between trajectory methods and occupancy grids, proposing a dynamic program for decoding likely trajectories from occupancy grids under a simple motion model. 
MotionNet \cite{wu2020motionnet} introduced a joint perception and motion prediction network that outputs BEV maps, with each grid cell encoding object category and motion information. In response to the demand for real-time prediction, Terwilliger \textit{et al.} \cite{Terwilliger2019} proposed semantic forecasting by predicting current frame segmentation and future optical flow with RNN-based networks. MP3 \cite{casas2021mp3} advances a concept akin to Flow Fields, termed Motion Fields, to predict forward motion vectors and associated probabilities per grid cell, facilitating planning tasks. Following this line, Fiery \cite{fiery2021} employed instance segmentation to predict future occupancy grids and motion flows of dynamic agents. 
Waymo \cite{mahjourian2022occupancy} presented a model employing backward flows to predict both occupancy and flow in a spatio-temporal grid. Additionally, Waymo launched a challenge \cite{Waymochallenge} addressing the same research problem. This challenge, however, considers availability of tracked agents as well as prior road map information.

Different from these methods, we employ probabilistic version of DOGMs \cite{lukas22} as our input source, which provide static and dynamic occupancy states of the scene in a BEV grid. Our input DOGMs not only offer probabilistic information even when only partial data is available, but also facilitate seamless integration of various sensor inputs and configurations without retraining the entire network.

%% file: sections/approach.tex
\section{System Overview} \label{sec:approach}
Our methodology builds upon the overall pipeline proposed in \cite{asghar2023vehicle}, which addresses vehicle motion forecasting into semantic grids from OGMs. This work specifically focuses on incorporating scene flow to further improve the vehicle motion predictions as semantic grids. 
We discuss in detail here our proposed approach, and the pipeline is summarized in Fig. \ref{fig:overview}.

\subsection{Problem Formulation}\label{sec:approach-b}
We formally define the task of vehicle motion prediction, see Fig. \ref{fig:overview}.
Let $X_t \in \mathrm{R}^{6 \times w \times h}$ be the $t$-th frame of the input grid where $w$ and $h$ denote the  width and height respectively. $X_t$ comprises of three occupancy state channels, two velocity channels, and one semantic grid channel. Let $Z_t \in \mathrm{R}^{3 \times w \times h}$ be the $t$-th frame of the output grid that contains a single vehicle semantic grid channel, ${Y}_{t}  \in \mathrm{R}^{1 \times w \times h}$, and two flow grid channels, ${F}_{t}  \in \mathrm{R}^{2 \times w \times h}$.

Given a set of input sequence $X_{t-N:t}$, the task of our proposed multi-head network is to predict 1) the vehicle grid $\hat{Y}_{t}$ at the current time step, and 2) the sequence of the future grids $\hat{Z}_{t+1:t+{P}}$ . This sequence encompasses both the future vehicle semantic grids $\hat{Y}_{t+1:t+{P}}$, and the future flow grids $\hat{F}_{t+1:t+{P}}$, with $P$ representing the prediction horizon.

Next, we utilize the predicted sequence of flow grids $\hat{F}_{t+1:t+{P}}$ to warp the vehicle grid $\hat{Y}_{t}$ recursively, and obtain an additional sequence of warped future vehicle semantic grids $\hat{W}_{t+1:t+{P}}$. This warping process can be expressed by Eq. \ref{eq:warp}, 

\begin{align}
\hat{W}_{t+1} =  f_w (\hat{W}_{t}, \hat{F}_{t+1} )  
\label{eq:warp}
\end{align}

where $f_w$ is the warping operation and $\hat{W}_{t} = \hat{Y}_{t}$ initially.

\subsection{Input Scene Representation}\label{sec:approach-a}
\subsubsection{Dynamic occupancy grid maps} DOGMs provide a grid-based representation of the environment in a bird's-eye view. Each cell in the grid is estimated in parallel within the system and contains information about its occupancy state, as well as associated dynamics.
Unlike classic object detection-and-tracking methods, such approaches mitigates risks linked to uncautious thresholding.

We incorporate Bayesian dynamic occupancy grid filter \cite{lukas22} to generate DOGMs from LiDAR sensor data. This approach is used to estimate probabilities associated with four possible occupancy states for each cell in the grid: i) free, ii) occupied and static, iii) occupied and dynamic, and iv) unknown occupancy. Additionally, the framework provides velocity estimates for each cell, represented by mixtures of grids and particle sets.

\begin{figure}[t]
	\centering
	    \includegraphics[trim={0 0 0 0},clip,width=1.0\columnwidth]{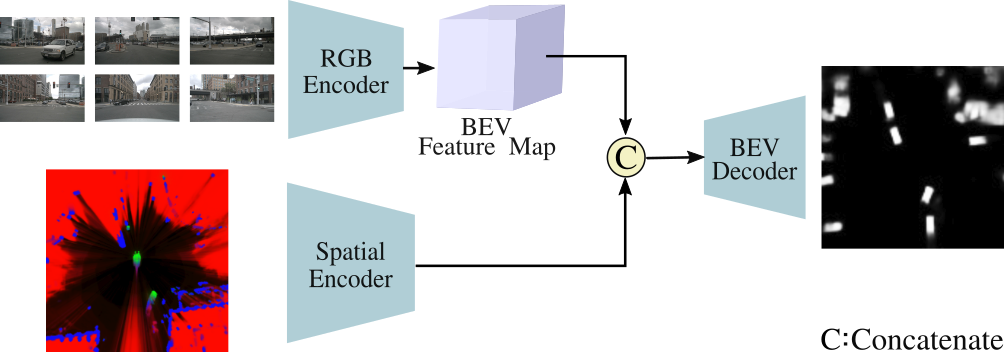}
\caption{\small To obtain semantic information, we fuse BEV features encoded from camera images with occupancy state grid. The Red, Green, and Blue channels in the RGB grid denote the unknown state, dynamic and static occupied states respectively. The black areas within the occupancy state grid indicate free space.}
\label{fig:semantics}
\vspace{-0.5cm}
\end{figure}

In the proposed network, the input $X_t \in \mathrm{R}^{6 \times w \times h}$ comprises six channels, with five channels sourced from the output of a DOGM. These encompass information pertaining to three DOGM occupancy states (the definitive visualization represented in Fig. \ref{fig:semantics}) and two-channel DOGM velocities along the $x$ and $y$-axis. 
Figure 1 showcases the DOGM state grids and the velocity grids depicted below as two-dimensional flow illustrations in the input.
An additional layer of semantic information is overlaid on top of the DOGM state grids, depicted in white. The inclusion of semantic information is discussed in the section below.

\subsubsection{Semantic information}\label{subsec:semantics}
A semantic grid contains the probability assigned to each cell for being occupied by a specific semantic category, such as vehicles in this case. 
The approach outlined in the Lift, Splat, Shoot (LSS) \cite{philion2020lift} is repurposed in this work to prepare the semantic information for input into our proposed network. 
LSS extracts feature maps from multiple input camera images in BEV and decodes them into vehicle occupancy prediction represented as a semantic grid. Our adaptation involves concatenating corresponding DOGMs alongside the feature maps extracted from camera images and decoding the combined information into semantic grids. This fusion of DOGMs with BEV features results in a more comprehensive understanding of the scene. This modified pipeline is illustrated in Fig \ref{fig:semantics}.

Note that the semantic information can be incorporated in this framework via alternative methods as well. For example, vehicle semantic information from camera images can be integrated into the Bayesian perception framework \cite{lukas22}, allowing for probabilistic estimation alongside the DOGM.

\subsection{Flow-guided Predictions}\label{sec:approach-a-b}
The network is equipped with two decoder heads: one is dedicated to predicting a semantic grid for current time step, while the other forecasts sequential future grids and flows simultaneously.

\begin{figure}[t]
	\centering
	    \includegraphics[trim={0 40 80 0},clip,width=0.8\columnwidth]{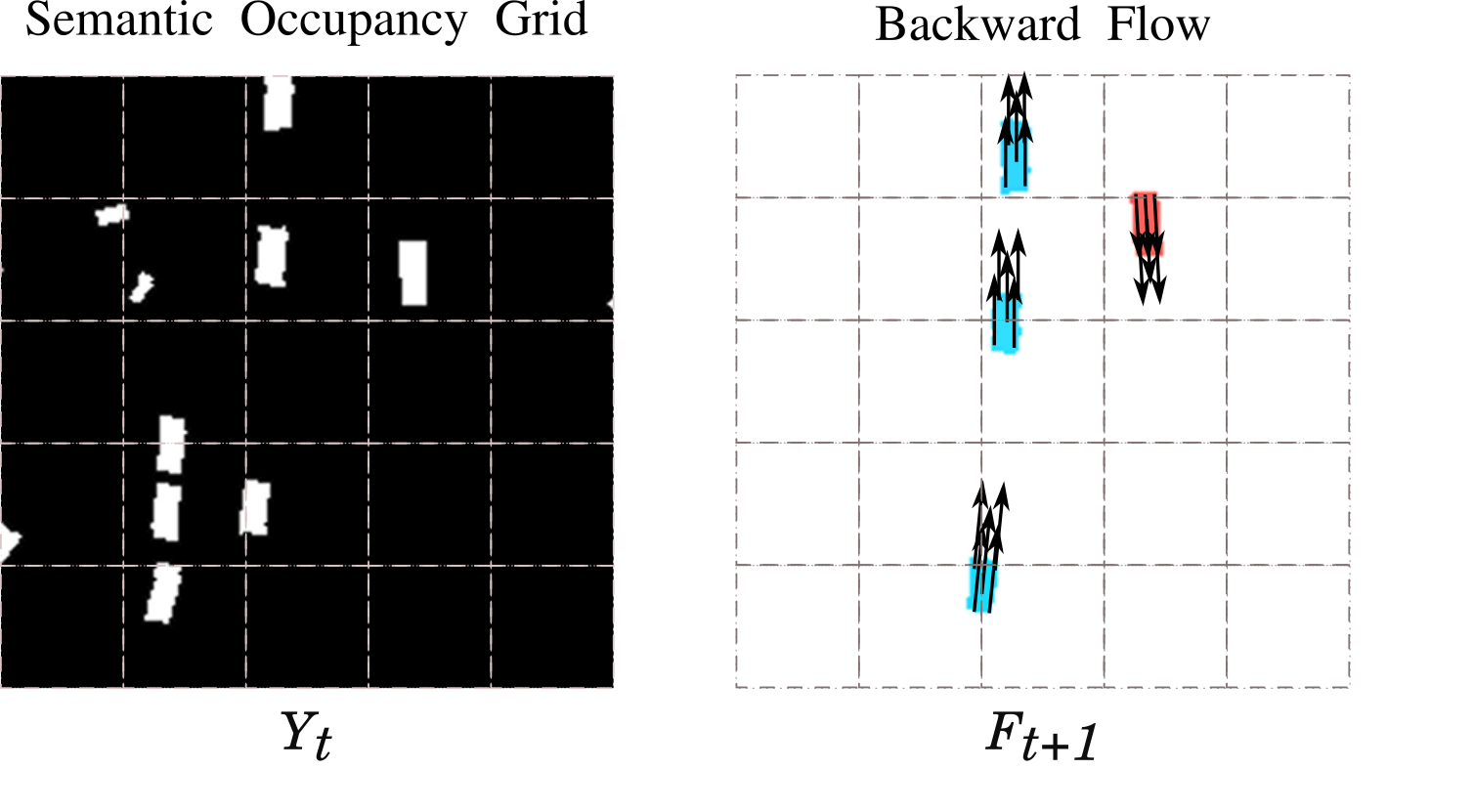}
\caption{\small In $F_{t+1}$, backward flow vectors indicate the motion of vehicles and point to the origin of their respective occupancy in $Y_t$. Red and blue represent dynamic vehicles moving upward and downward within the grid, respectively.}
\label{fig:bck-flow}
\vspace{-0.5cm}
\end{figure}

The flow is represented as a two-channel grid that models the velocity of vehicles along the $x$ and $y$ dimensions.
We utilize backward flow to capture vehicle motion from their future positions back to their current positions.
At each time step, the backward flow contains velocity vectors for each cell, indicating the change in occupancy from the previous time step. 
Refer to Fig. \ref{fig:bck-flow} for a visual illustration.

To generate a sequence of flow-guided semantic grid, we start with the current semantic grid $\hat{Y}_{t}$. This grid is initially warped using the flow $\hat{F}_{t+1}$ to obtain a warped semantic grid $\hat{W}_{t+1}$. Subsequently, $\hat{W}_{t+1}$ is further warped with the next flow $\hat{F}_{t+2}$. This process is repeated recursively for all the subsequent time steps to yield a sequence of warped vehicle semantic grids.
Worth noting, our method not only predicts the flow of occupancy but also incorporates the potential flow in free spaces.

Alongside predicting the flow, the network also forecasts a sequence of future semantic grids. This simultaneous prediction aids the network in correlating the occupancy of the semantic grid with the corresponding flow vectors.

\subsection{Model Architecture}\label{subsec:NN-architectures}

The model combines spatio-temporal predictions with a conditional variational approach.

\subsubsection{Endoder} Each frame in the input sequence, denoted as $X_{t-N:t}$,  is passed through the spatial encoder. Then the generated features are sequentially fed into a ConvLSTM (Convolutional Long Short Term Memory) block, enabling the simultaneous capture of spatial and temporal information.  To achieve this, we use 4 ConvLSTM units, each with 128 hidden features, ensuring the extraction of spatio-temporal features that encompass the past evolution of the scene.

\subsubsection{Probablistic modeling}\label{subsubsec:probab} To allow for multimodal predictions, we adopt a conditional variational approach \cite{hu2020probabilistic}. This approach allows the network to learn two distributions that capture the evolution of the scene in latent space: i) present distribution that is conditioned on present spatiotemporal features and ii) future distribution that is conditioned on present spatiotemporal as well as future ground truth. Both distributions are parameterized as diagonal Gaussian distributions, each consisting of 32 latent space dimensions.

\subsubsection{Decoder} The spatial GRUs utilize spatio-temporal features to iteratively forecast future states, conditioned on samples from Gaussian distributions. Each timestep comprises of 3 units of convolutional GRUs.
These multi-step future states are then fed to the two decoder heads, as discussed in \ref{sec:approach-a-b}.

 For more details on the model, we refer the reader to \cite{asghar2023vehicle}.

\subsection{Losses}\label{subsec:NN-losses}
The network is trained to learn vehicle motion forecasting as semantic grids and their corresponding flow representations. All vehicle semantic grids are trained with binary cross-entropy loss $\mathcal{L}$$_{BCE}$ with a positive sample weight of 5. This includes loss for $\hat{Y}_{t}$, $\hat{Y}_{t+1:t+{P}}$, and $\hat{W}_{t+1:t+{P}}$. The loss for warped semantic grids $\hat{W}$ acts as self-supervised loss for the flow grid $\hat{F}$.
\vspace{+0.05cm}
\begin{align*}
\mathcal{L}_{{BCE}_{sum}} = \lambda_{d} \mathcal{L}_{{BCE}_{\hat{Y}_{t}}} +  \lambda_{b} \mathcal{L}_{{BCE}_{\hat{Y}_{t+1:t+{P}}}} + \lambda_{w} \mathcal{L}_{{BCE}_{\hat{W}_{t+1:t+{P}}}}
\end{align*} \label{eq:loss1}
\vspace{+0.05cm}
where $\lambda_{d}$, $\lambda_{b}$ and $\lambda_{w}$ are the loss weights for  $\hat{Y}_{t}$, $\hat{Y}_{t+1:t+{P}}$, and $\hat{W}_{t+1:t+{P}}$ respectively. 

For the supervised loss of the flow grid $\hat{F}$, we define an L1 loss, only for the cells that are occupied by the vehicles.
Additionally, we introduce Kullback-Leibler divergence loss $\mathcal{L}$$_{KL}$ to encourage alignment between the present and the future distributions, described in \ref{subsubsec:probab}.

The final loss is weighted sum of all these losses.
\vspace{+0.05cm}
\begin{align*}
\mathcal{L} = \mathcal{L}_{{BCE}_{sum}} + \lambda_{f} \mathcal{L}_{flow} +  \lambda_{k} \mathcal{L}_{KL}    
\end{align*} \label{eq:loss2}
\vspace{+0.1cm}
where  $\lambda_{f}$  and $\lambda_{k}$ are the respective loss weights. 

%% file: sections/experiments.tex
\section{Experiments} \label{sec:implementation}

\begin{figure*}[ht]
	\centering
	    \includegraphics[trim={0 0 0 0},clip,width=1.69\columnwidth]{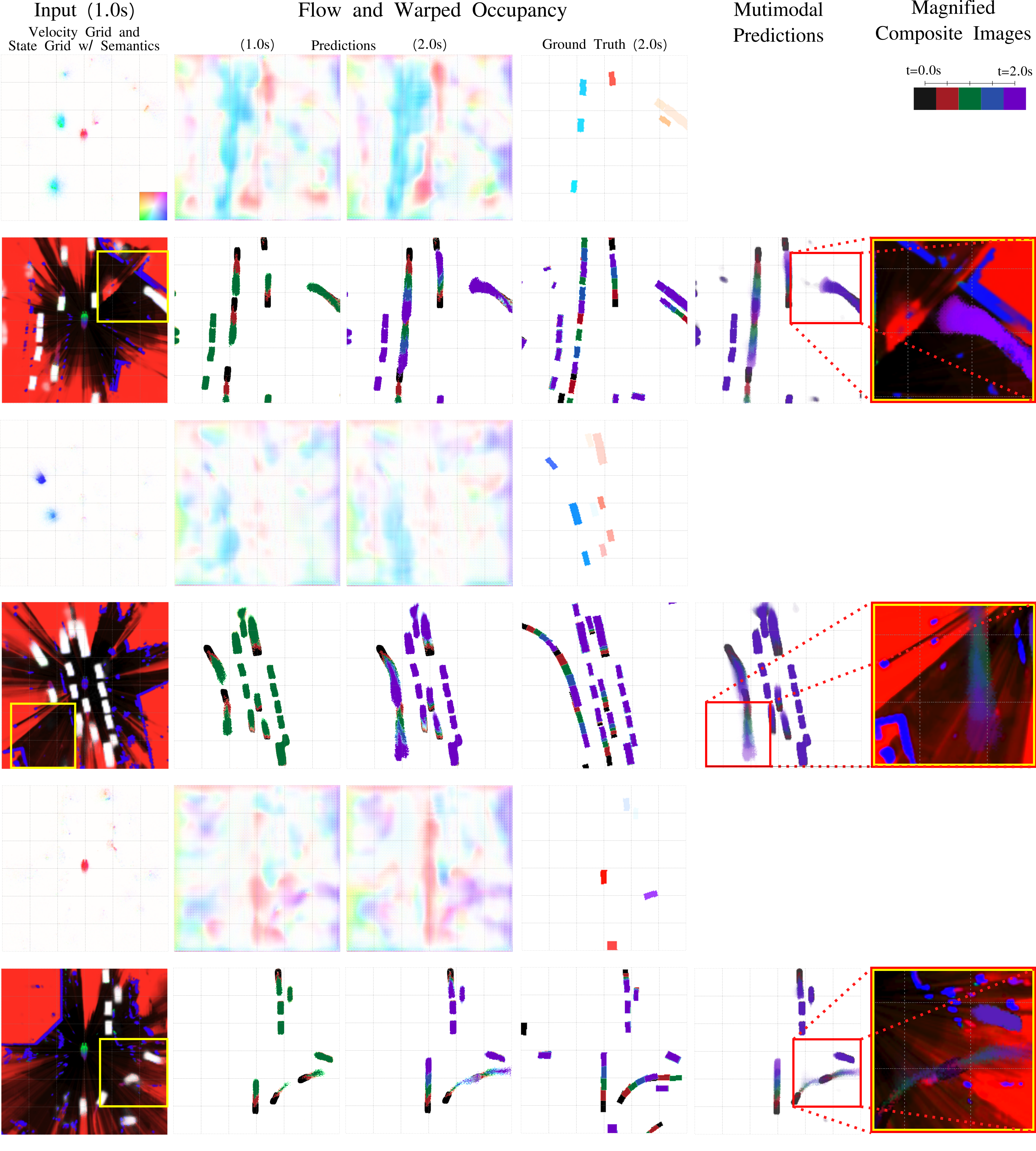}
\caption{\small Semantic flow and warped occupancy prediction examples are demonstrated on three scenes from the Nuscenes Datatset \cite{nuscenes2019}, covering an area of 60x60m. The first cloumn displays the DOGM input with semantic labels (in white) at the latest timestep. Flow and warped occupancies are shown for 1.0s and 2.0s. Occupancies are color-coded to visualize the range from 0.0s to 2.0s, transitioning from black to purple. The last column showcases magnified current static part of the scene, on which are displayed agent predictions. The magnified scene correspond to the yellow and red boxes, providing a zoomed-in view of the multimodal predictions overlaid on occupancies in the DOGM.}\label{fig:qualitative} 
\vspace{-0.5cm}
\end{figure*}

\subsection{Dataset}\label{subsec:experiments}
We evaluate our proposed model on the real-world NuScenes dataset \cite{nuscenes2019}. The original dataset consists of 700 and 150 scenes for training and validation respectively. Each scene has a duration of \SI{20}{\second}, with lidar data available at 10Hz and annotated keyframes at frequency of 2Hz. The lidar based DOGMs are generated comprising of state grids and velocity grids, with a resolution of \SI{0.1}{\metre}. For semantic grid prediction, all 6 camera images at each keyframes are considered along with respective DOGM state grids.

The ground truth annotations for semantic grids are prepared based on the vehicle bounding box annotations provided in the dataset. 
The velocity or speed of the vehicles is unavailable. 
Therefore, the ground truth flow of the vehicles is estimated by calculating the displacement of vehicle centroids on the grid, which is then normalized by the grid dimensions.

All grids are initially generated ego-centrically, with the ego-vehicle positioned at the center and facing upward.
Future predictions in each sequence are made relative to the ego-vehicle frame of reference at the latest timestep.
The past and future grids are subsequently transformed to facilitate allo-centric predictions in a fixed frame, using the available odometry data from the dataset, and then cropped to cover an area of 60x60m.

In total, we have 23K training and 5K validation sequences respectively.

\subsection{Training}\label{subsec:training}
The input sequence $X_{t-{2}:t}$ consists of 3 frames, spanning over 1.0s. The network is trained to make predictions for present $\hat{Y}_{t}$ and 4 future frames $\hat{Y}_{t+1:t+{4}}$, for instances 0.5s to 2.0s, with steps of 0.5s.
For training, the grid images are resized to 240x240 pixels, thus each pixel has a resolution of 0.25m.
The loss coefficients are set to  $\lambda_{b}=1.0$,  $\lambda_{w} = 1.0$ , $\lambda_{f} = 0.05 $ and  $\lambda_{k} = 0.005$. 
Our network is implemented on Pytorch and trained on 8  Nvidia Tesla V100 GPUs. 
Adam Optimizer is employed at the learning rate of $3$x${10}^{-4}$ and weight decay of $1$x${10}^{-7}$. Dataset is trained for 20 epochs with the batch size of 18. This takes approximately 5 hours.

%% file: sections/results.tex
\section{Results} \label{sec:results}

\subsection{Qualitative Evaluation}\label{subsec:qltyevaluation}
Figure \ref{fig:qualitative} shows predicted flow and warped occupancies at \SI{1.0}{\second} and \SI{2.0}{\second} across three scenes.
The two-dimensional flow vectors are represented as RGB images, with colors indicating various vector directions, while occupancies are depicted with distinct colors at each timestep. 
Multimodal predictions are generated using 100 random samples from a Gaussian distribution, with color intensity corresponding to the frequency of occupancy at each cell. 
Additionally, semantic channels are superimposed on the state grids in the input for visualization purposes. Notably, the network learns scene structure from DOGM occupancy information and the behavior of dynamic vehicles, as map or road information is unavailable in the input.

The network's flow grid predictions focus on warping the semantic grid from one timestep to the next, rather than specifically capturing the flow of occupancy. This allows for predicting both the movement of dynamic vehicles and the resulting free space. In the flow images, white spaces dominate the static scene structure observed in the input state grids. Regions with no observed occupancies, whether in the input or the predicted horizon, contain random values in the flow grid as they do not impact the predicted grids.

\subsubsection*{\textbf{Collision Avoidance}}\label{subsec:collision}
A notable attribute in the flow-warped semantic grids is their effective collision avoidance capability during predictions. 
The motion prediction of dynamic vehicles consistently demonstrates a tendency to navigate away from static occupancies within the scene, regardless of whether they are static vehicles or unrecognized categories. As illustrated in Fig. \ref{fig:qualitative}, multimodal predictions avoid the static obstacles within the scene, as highlighted in the magnified sections of grids in the rightmost column. For instance, in the first example, bus predictions exhibit cautious behavior in the face of static cells ahead. Similarly, in the third scenario, the multimodal predictions steer clear of small static obstacles within the space.

Furthemore, the predictions in dynamic spaces also show notable improvement. The semantic grid predictions without flow tend to get blurry when multiple dynamic vehicles are in close proximity due to uncertainty, for examples refer to \cite{asghar2023vehicle}. However, with the inclusion of flow, predictions become sharper, resulting in clearer delineation even in dense traffic. This enhancement is particularly evident in the second example, especially among the top vehicles in the scene.

\subsection{Quantitative Evaluation}\label{subsec:qntyevaluation}

\subsubsection{Ablation Study}\label{subsec:ablationevaluation}
We conduct an ablation study with different configurations of the proposed network to investigate the role of semantic flow. We consider 3 different variations for the future prediction decoder: i) only predict semantic grids  $\hat{Y}_{t+1:t+{P}}$ , ii) only predict flows $\hat{F}_{t+1:t+{P}}$ , and iii) predict both together. These variations are trained with and without the velocity grid in the input.

We evaluate the performance of the semantic and warped grids using binary classification evaluation metrics: Intersection of Union (IoU) and Area under the precision-recall curve (AUC). IoU is computed with a threshold of 0.5, while AUC is calculated across 100 linearly spaced thresholds.
Table \ref{table:ablation-study} shows the average IoU and AUC values over a 2.0s prediction horizon for six cases. 
The IoUs for our proposed architecture are similar to those of the method solely predicting semantic grids, see the first and the last row. 
However, incorporating flow predictions significantly enhances AUC, indicating improved precision and recall - the proportion of correct vehicle predictions and correctly predicted occupancies, respectively.

The integration of DOGM velocity grid and prediction of semantics and flow
collectively result in optimal IoU and AUC, enhancing the network capacity for behavior learning.
Interestingly, employing the same configuration without velocity input yields inferior scores, suggesting the crucial contribution of velocity inclusion in avoiding local minima during training.

\begin{table}[h]
\vspace{-0.2cm}
\caption{\small Ablation study on input and output configuration of the architecture. We report mean values of IoU and AUC over a 2.0s prediction horizon. Bold indicates best.}
\centering
\begin{tabular}{cccccc}
\cline{1-3}
\hline
\multicolumn{1}{c}{\textbf{Input}} & \multicolumn{1}{c}{\textbf{Output}} & \multicolumn{2}{c}{\textbf{IoU}$(\uparrow)$}  & \multicolumn{2}{c}{\textbf{AUC}$(\uparrow)$}\\
\hline
\multicolumn{1}{c}{\textbf{Velocity}} & \multicolumn{1}{c}{\textbf{Semantics/Flow}} & \multicolumn{1}{c}{\textbf{Sem}}   & \multicolumn{1}{c}{\textbf{Warped}}  & \multicolumn{1}{c}{\textbf{Sem}}   & \multicolumn{1}{c}{\textbf{Warped}}\\
\cline{1-6} 
\multicolumn{1}{c}{} & {Semantics only} & {0.493} & {-} & {0.613}  & {-} \\
\multicolumn{1}{c}{\Checkmark} & {Semantics only} & {0.487} & {-}  & {0.626}  & {-} \\
\multicolumn{1}{c}{} & {Flow only} & {-} & {0.481} & {-}   & \textbf{0.631} \\
\multicolumn{1}{c}{{\Checkmark}} & {Flow only} & {-} & {0.480}  & {-}  & {0.630} \\
\multicolumn{1}{c}{} & {Semantics + Flow} & {0.485} & {0.463} & {{0.629}}   & {0.613} \\
\multicolumn{1}{c}{{\Checkmark}} & {Semantics + Flow} & \textbf{{0.503}} & \textbf{{0.493}}  & \textbf{{0.649}}  & \textbf{0.631}\\ \hline
\end{tabular}
\label{table:ablation-study}
\end{table}

\subsubsection{Retention of vehicles}\label{subsec:OGMevaluation}

We further evaluate the network ability to predict static and dynamic agents by tracking individual agents. In our observations, dynamic agent predictions tend to lose clarity with each time step and may eventually become indistinct. Similarly, predictions of vehicles in semantic grids, though improved, also gets blurrier or even disappears when the network is uncertain of vehicle behavior.

To assess network's ability to retain vehicles in the scene, we track vehicles perceived in the current semantic grid $\hat{Y}_t$ individually. If the prediction of the vehicle does not overlap with the ground truth semantic grid, we consider it \textit{lost}. We define overlap as at least 10 cells with an occupancy probability higher than 0.3, covering an area of \SI{0.625}{m^2} (approximately a third of a vehicle).
We study three cases: a) using four-channel semantic-OGM predictions with a video prediction network PredRNN \cite{wang2022predrnn}, b) our proposed network without flow, and c) the complete proposed network.

\begin{table}[h]
\vspace{-0.2cm}
\caption{\small Retention of dynamic and static vehicles lost in a 2.0s prediction horizon. Bold indicates best.} 
\centering
\begin{tabular}{c|c|c|c|c|c|}
\cline{1-5} 
\hline
\multicolumn{1}{c}{\textbf{Network}} & \multicolumn{2}{c}{\textbf{ Vehicles Retained (\%)}$(\uparrow)$}  & \multicolumn{2}{c}{\textbf{}}\\\hline
\cline{1-5} 
\multicolumn{1}{c}{\textbf{}} & \multicolumn{1}{c}{\textbf{Dynamic}}   & \multicolumn{1}{c}{\textbf{Static}}  & \multicolumn{2}{c}{}\\
\cline{2-3} 
\multicolumn{1}{c}{PredRNN \cite{wang2022predrnn}} & \multicolumn{1}{c}{{90.2}}  & \multicolumn{1}{c}{97.0}  & \multicolumn{2}{c}{{}} \\
\multicolumn{1}{c}{Ours w/o flow} & \multicolumn{1}{c}{{95.8}} & \multicolumn{1}{c}{99.4}  & \multicolumn{2}{c}{} \\ 
\multicolumn{1}{c}{Ours}  & \multicolumn{1}{c}{\textbf{97.6}} & \multicolumn{1}{c}{\textbf{99.7}}  & \multicolumn{2}{c}{\textbf{}} \\ \hline
\end{tabular}
\label{table:results-b}
\end{table}

We show the percentage of static and dynamic vehicles retained by the three methods in Table \ref{table:results-b}, considering a total of 10K dynamic and 20K static vehicles in the validation sequences. 
Under these metrics, our method tends to retain more agents, static or dynamic, than mere classic video prediction approach, with the inclusion of flow information further improving the results.

%% file: sections/conclusion.tex
\section{Conclusion}\label{sec:conclusion}
In this work, we introduce a novel forecasting approach that leverages DOGMs and semantic information to predict scene evolutions as flow-guided semantic grids. Our framework is capable of multi-modal predictions, learning patterns and behaviours based on the observable static scene and motion of dynamic agents. 
By predicting flow alongside semantic grids, this methods tends to improve prediction results, especially in dynamic parts.
While only vehicles were considered in the flow-guided agent prediction, such method should in the future be expanded to the prediction of other types of agents, and unidentified grid-level components.